\documentclass{article} 
\usepackage{iclr2024_conference,times}

\usepackage{amsmath,amsfonts,bm}

\usepackage{hyperref}
\usepackage{url}
\usepackage{inconsolata}

\usepackage{tikz}
\usepackage{pgfplots}
\usepackage{xspace}
\usepackage[frozencache,cachedir=.]{minted}

\usepackage[T1]{fontenc}

\usepackage[utf8]{inputenc}

\usepackage{microtype}
\usepackage{makecell}

\usepackage{array}
\usepackage{booktabs}
\usepackage{soul}
\usepackage{url}
\usepackage[utf8]{inputenc}
\usepackage{graphicx}
\usepackage{amsmath}
\usepackage{booktabs}
\usepackage{colortbl}
\usepackage{multirow}
\usepackage{float}	
\usepackage{dblfloatfix}
\usepackage{setspace}  

\usepackage{xcolor}	

\usepackage{cleveref}

\definecolor{ForestGreen}{rgb}{0.13, 0.55, 0.13}
\definecolor{skyblue}{HTML}{46C5DD}	

\newcommand{\eat}[1]{}

\newcommand{\gapx}{\hspace*{2mm}}

\usepackage{quoting}
\newenvironment{ite}{                     
     \parskip 0cm \begin{itemize} \parskip 0cm \parsep 0cm \itemsep 0cm \topsep 0cm}{
        \end{itemize}} 
\usepackage{algorithm}
\usepackage[noend]{algpseudocode}

\usepackage[]{mdframed}
\usepackage{framed}

\usepackage{ulem}

\usepackage{ctable} 
\usepackage{multicol}

\renewcommand{\cite}{\citep}

\definecolor{lightgray}{gray}{0.9}
\definecolor{Box1Color}{RGB}{227, 236, 246}
\definecolor{Box2Color}{RGB}{248, 220, 225}
\definecolor{Box3Color}{RGB}{255, 238, 224}
\definecolor{cbBlue}{RGB}{0, 114, 178}
\definecolor{cbOrange}{RGB}{240, 228, 66}
\definecolor{cbGreen}{RGB}{0, 158, 115}
\definecolor{cbRed}{RGB}{213, 94, 0}
\definecolor{cbPurple}{RGB}{204, 121, 167}
\definecolor{cbSkyBlue}{RGB}{86, 180, 233}
\definecolor{cbGray}{RGB}{128, 128, 128}
\definecolor{CBF1}{RGB}{255,99,132}  
\definecolor{CBF2}{RGB}{54,162,235}  
\definecolor{CBF3}{RGB}{255,206,86}  
\definecolor{CBF4}{RGB}{75,192,192}  
\definecolor{CBF5}{RGB}{153,102,255} 
\definecolor{CBF1b}{RGB}{205,89,112}  
\definecolor{CBF2b}{RGB}{44,142,215}  
\definecolor{CBF5b}{RGB}{133,92,225}  
\definecolor{PromptBgColor}{RGB}{255, 238, 224}

 \title{CLIN: Continual Learning in Language Agents for Rapid Task Adaptation and Generalization}

\title{CLIN: A goal-driven agent that continually improves by learning causal abstractions}

\title{CLIN: A Continually Learning Language Agent for Rapid Task Adaptation and Generalization}

\author{Bodhisattwa Prasad Majumder\textsuperscript{$1$}, Bhavana Dalvi Mishra\textsuperscript{$1$},\\
\textbf{Peter Jansen\textsuperscript{$1,2$}, Oyvind Tafjord\textsuperscript{$1$}, Niket Tandon\textsuperscript{$1$}, Li Zhang\textsuperscript{$3$},}\\
\textbf{Chris-Callison Burch\textsuperscript{$3$}, Peter Clark\textsuperscript{$1$}}
\\
\textsuperscript{$1$}Allen Institute of AI\\
\textsuperscript{$2$}University of Arizona\\
\textsuperscript{$3$}University of Pennsylvania\\ \\
Contact: \texttt{\{bodhisattwam, bhavanad\}@allenai.org}\\
Project page: \href{https://allenai.github.io/clin/}{\texttt{https://allenai.github.io/clin/}}
}

\newcommand{\clin}{\textsc{CLIN}}

\newcommand{\genenv}{\textsc{Gen-Env}}
\newcommand{\gentask}{\textsc{Gen-Task}}
\newcommand{\genadapt}{\textsc{Gen-Adapt}}

\iclrfinalcopy 
\begin{document}
\maketitle

\begin{abstract}

Language agents have shown some ability to interact with an external environment, e.g., a virtual
world such as ScienceWorld, to perform complex tasks, e.g.,  growing a plant, without the startup
costs of reinforcement learning. However, despite their zero-shot capabilities, these agents
to date do not continually improve over time, beyond performance refinement on a specific task.
 Here we present CLIN, the first language-based agent to achieve this,
so that it continually improves over multiple trials, including when both the
environment and task are varied, and without requiring parameter updates.
Our approach is to use a persistent, dynamic, textual memory,
centered on {\it causal abstractions} (rather than general ``helpful hints''),
that is regularly updated after each trial so that the agent gradually learns useful
knowledge for new trials. In the ScienceWorld benchmark, CLIN is able to continually improve on repeated trials on the same task and environment,
outperforming state-of-the-art reflective language agents like Reflexion by 23 absolute points.
CLIN can also transfer its learning to new environments (or new tasks), improving its zero-shot performance
by 4 points (13 for new tasks) and can further improve performance there through continual memory updates, enhancing
performance by an additional 17 points (7 for new tasks). 
This suggests a new architecture for agents built on frozen models that can still 
continually and rapidly improve over time.

\end{abstract}

\section{Introduction}

Large language models (LLMs) have been increasingly used to interact with external
environments (e.g., simulated worlds) as goal-driven agents \cite{Reed2022AGA}. However, it has been
challenging for these language agents to efficiently learn from trial-and-error
as traditional reinforcement learning methods require extensive training samples and
expensive model fine-tuning \cite{Chen2021DecisionTR, Ammanabrolu2020HowTM}. More recently, new techniques have appeared in which
an agent reflects on its own past experience solving a task in a particular environment,
and generates language-based insights to help it retry the task, e.g., Reflexion \cite{reflexion}.
Such methods have the advantage of not requiring parameter updates (particularly useful given
the growing popularity of frozen language models). However, the style of such insights plays
a crucial role in performance, and not all insights improve generalization performance. For example,
a specific insight such as "In the next trial, I will go to desk 1 and find the lamp"
\cite{reflexion} may have limited value (or even hurt) for a different environment
or task.

\begin{figure}
    \centering
    \includegraphics[trim= 75 300 125 85,clip, width=0.95\textwidth]{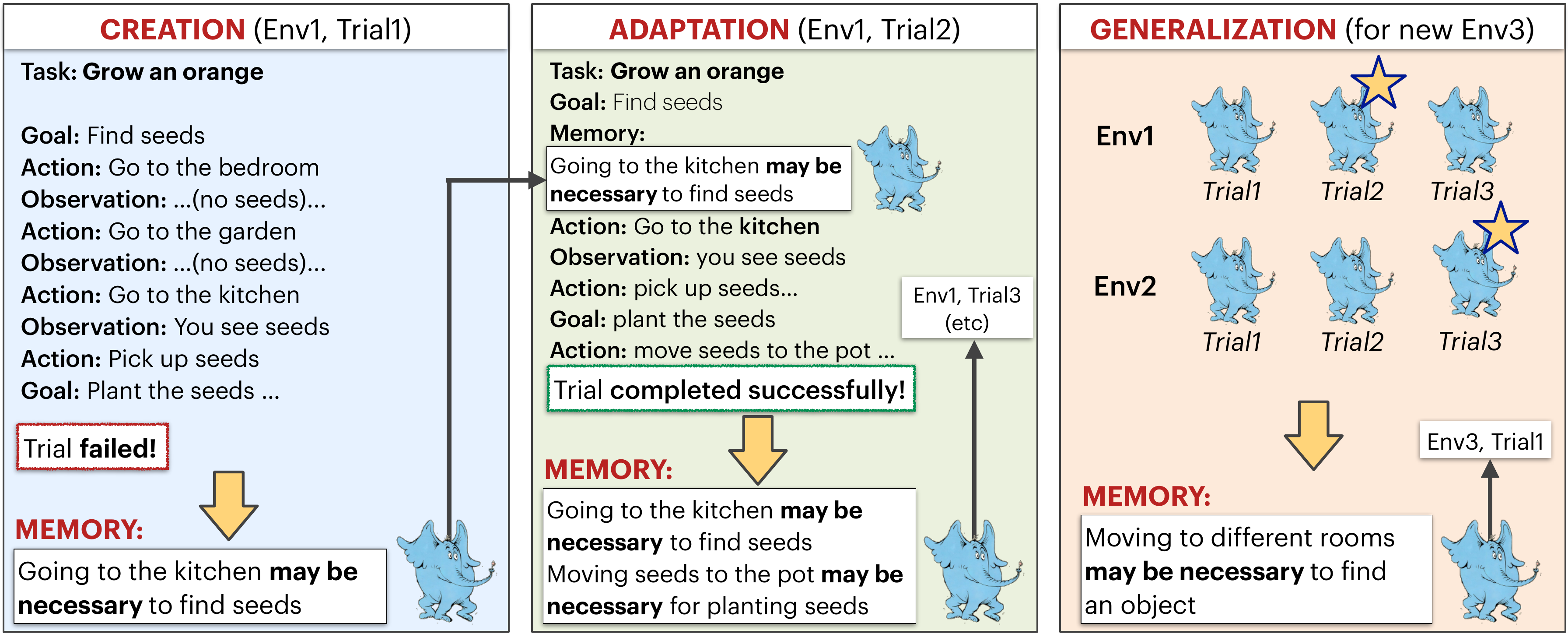}
    \caption{CLIN creates (Trial1) or adapts (Trial2+) a {\bf memory of causal abstractions} to help in future trials by reflecting on the last trial and current memory. It does this using a suitably prompted LLM to generate the updated memory (Section~\ref{sec:memory}). Here, reflecting on Trial1, CLIN notes in memory that going to the kitchen helped with finding seeds, enabling it to find the seeds faster in Trial2. From there, it also learns that moving the seeds to the pot helped plant the seeds. To further generalize across episodes (sequences of trials, right figure) for use in new environments, CLIN generates a summary (“meta-memory”) of the best (starred) memories from each prior episode, here generating the generalization that moving to different rooms helps finding objects (Section~\ref{sec:metamemory})
    \label{fig:enter-label}}
\end{figure}


Our goal is a system that will continually improve over time, both while attempting
the same task in the same environment, and across different tasks and environments.
Our approach builds on prior work on reflection in two ways: First, we conjecture that a
specific {\it style} of insight will be useful, namely one that captures {\bf causal abstractions}
about agent's actions, e.g., ``opening doors may be necessary for
movement between rooms''. Causal abstractions can potentially help the agent decide
which action to take in the future, and can be viewed as a kind of action model
learning \cite{Arora2018ARO}, but placed in the modern context of language models.
Second, we maintain these abstractions in a {\bf continually
evolving, dynamic memory}, which is regularly updated as the agent gains experience,
allowing useful causal knowledge to persist (and unhelpful knowledge to be dropped)
over time and between tasks and environments, as illustrated in Figure~\ref{fig:enter-label}.

We operationalize and evaluate this approach in a memory-augmented language agent called
CLIN (\textbf{c}ontinual \textbf{l}earning from \textbf{in}teractions). CLIN is an agent that operates in
ScienceWorld \cite{Wang2022ScienceWorldIY}, a virtual, text-based environment in which an agent is
tasked with science-based goals, e.g., boiling a liquid, growing a plant. We find that 
CLIN is able to rapidly learn about the environment and its action vocabulary and continually improve on repeated trials on the same task and environment,
outperforming state-of-the-art (SOTA) reflective language agents like Reflexion by 23 points.
CLIN can also transfer its learning to new environments (or tasks), improving its zero-shot performance
by 4 (13 for new tasks) points and can further improve performance through continual memory updates, enhancing
performance by an additional 17 (7 for new tasks) points. Our contributions are as follows:
\begin{ite}
\item For memory-based language agents, we show that memory of
  causal abstractions is effective at helping the agents learn over an extended
  period and in varying conditions.
\item We describe and evaluate CLIN, an architecture for a novel nonparametric learning paradigm. We
 show that CLIN learns faster than prior systems and generalizes better
 to new tasks and new environments, achieving state-of-the-art.
 \end{ite}
Overall, this work suggests that a dynamically maintained memory, centered
around causal knowledge, is a promising way forward for agents built
on frozen models to continually improve over time.

\section{Related Work \label{related-work}}

There is a long literature of work on agents that can navigate complex environments.
A common approach is to use reinforcement learning (RL), e.g., DRRN \cite{He2015DeepRL},
KG-A2C \cite{Ammanabrolu2020GraphCR},
CALM \cite{Yao2020KeepCA},
where agents learn a task over repeated trials. However, while effective,
such agents typically require a large number of trials to learn and have trouble
adapting to unexpected changes in the test environment. More recently,
\cite{Team2023HumanTimescaleAI} demonstrated AdA, an agent that could rapidly adapt to open-ended
novel 3D problems, using meta-reinforcement learning, essentially being able
to change its policy on the fly. However, AdA required vast amounts of
pretraining, and this skill was still limited to the style of environments and
problems seen in pretraining.

Recently, LLMs have provided a new tool for building
goal-directed agents \cite{llm-zero-shot-planner}. Given a linguistic description of the world state,
a task, and a history, the LLM can be prompted to suggest next actions
to take to achieve a goal, exploiting their wealth of semantic knowledge
about the world and requiring little training, e.g., SayCan \cite{Ahn2022DoAI},
ReAct \cite{Yao2022ReActSR}, and more recently SwiftSage \cite{Lin2023SwiftSageAG}, which
combines a supervised agent and a deliberative agent together. However, while
performing reasonably with little training data, such agents are unable
to learn and adapt from experience.

Two recent systems have demonstrated how a frozen-model-based agent could
improve at a task. Voyager \cite{Wang2023VoyagerAO} operates in the world of Minecraft,
growing a (code-based) skill library from rich feedback of its failures.
Reflexion \cite{reflexion} improves at a task by {\it reflecting}
on a failed attempt at that task and devise a new plan that accounted for
that mistake, used in the subsequent prompt to retry the task. While Reflexion did not have a long-term memory,
and its reflections were task- and environment-specific,
e.g., ``In the next trial, I will go to desk 1 and find the lamp.'', we take
inspiration from it to build an agent, \clin{}, which continually maintains and
adapts a long-term, persistent memory of reflections, useful across
different trials, tasks, and environments.

More generally, others have found that a memory of useful learnings can be
used to improve frozen LLM behavior, e.g., in QA
\cite{teachme,prompt-editing,Madaan2023SelfRefineIR}, or for modeling social behavior
\cite{simulacra}. We apply this finding to
goal-directed agents.

Finally, we note that the {\it content} of experiential memory is also important. Specifically, \clin{} learns a memory of {\it causal abstractions}, which
can be seen as learning a linguistic form of action model, describing the causal
effects of actions. While there has been substantial work in the 
planning community of learning action models in a fully formalized context
\cite{Arora2018ARO,Aineto2018LearningSA},
\clin{} loosely applies this idea in the linguistic world of LLM agents.

\section{Approach}

\textbf{Problem Formulation.}\hspace{1em}
Sequential decision-making tasks require agents to repeatedly observe and then act in an environment until they accomplish a specific goal.  At a high level, this can be accomplished by developing beliefs about the world, acting on the environment based on those beliefs, and then updating one's beliefs based on the observed outcome. Here, we investigate constructing an agent that can continually update its beliefs through interaction and observation while exploiting its past experience toward solving unseen parametric variations of tasks.

\textbf{Setup.}\hspace{1em} We investigate our continual learning agents in simulated environments. Our environments are modeled in a high-fidelity text-based simulator \cite{Wang2022ScienceWorldIY}, where both actions and observations are expressed in natural language. Let's define the task space to be $\mathcal{M}$, a collection of partially observable Markov Decision Processes (POMDPs) that can be executed in a set of environment configurations $\mathcal{E}$. Each task $m \in \mathcal{M}$ has an initial state and a desired winning state, which vary depending on the environment $e \in \mathcal{E}$. 

Our setup allows an agent to attempt a task several times; each time is denoted as a {\it trial}, $\mathcal{T}$, which consists of a total of $\tau$ steps. Each step comprises an action by the agent ($a$), and in response, the simulator returns the result of that action in the form of an observation ($o$) and a reward ($r$). A collection of $K$ trials is called an {\it episode}. The environment gets reset when the task reaches an end state (such as completing, failing, or timing out). In our continual learning setup, the agent retains its memory across trials/episodes, reaping the benefits of continued interaction with the environment.

\subsection{\textbf{\clin}: A Generative Agent \textbf{C}ontinually \textbf{L}earning from \textbf{IN}teractions}

\algrenewcommand\algorithmicindent{1.0em}%

\begin{figure}[t]
    \centering
    \includegraphics[trim= 290 400 185 155,clip, width=0.95\textwidth]{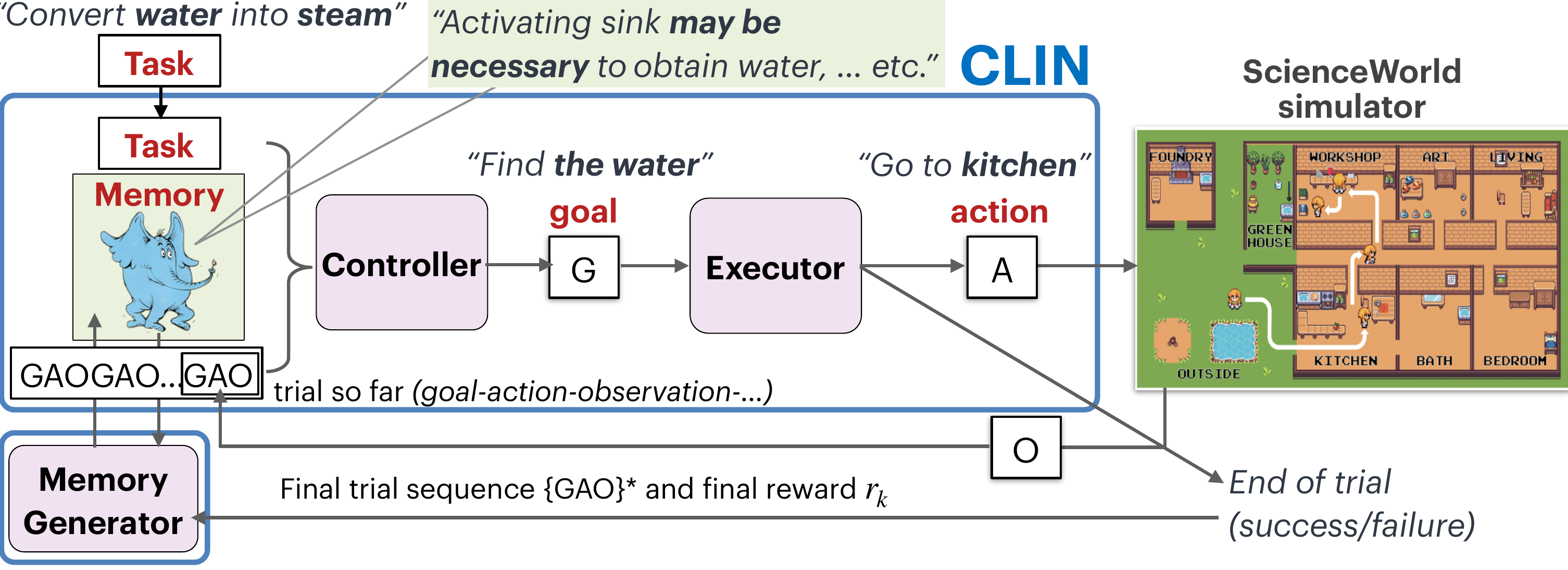}
    \caption{The architecture of CLIN. A {\bf controller} takes the current task, retrievals from memory, and the trial so far, to generate the next goal to achieve. The {\bf executor} then converts this to a valid action to perform towards that goal. The simulator then performs the action and returns an observation of that action's effect. Memory is updated at the end of each trial by the {\bf memory generator} (Section~\ref{sec:memory}).
    \label{architecture}}
    \vspace{-1mm}
\end{figure}

To act in the world, CLIN uses three modules: a {\bf memory}, a {\bf controller}, and an {\bf executor}, illustrated in Figure~\ref{architecture} and which we now describe.
Learning then occurs using a fourth module, a {\bf memory generator}, to generate an updated memory after each trial.

\textbf{Memory.}\hspace{0.7em} CLIN's memory ($\mathcal{S}$) is a persistent, dynamic collection of NL sentences that express causal abstractions, generated by reflecting on past experiences in order to help the agent perform better in the future. This generation process is described shortly in Section~\ref{sec:memory}. 
For example, having an explicit causal insight, ``opening the fridge is \texttt{necessary} to access apple juice'', learned from past experiences, can reduce the action search space for \clin~while looking for ``apple juice'' in the same environment in future trials. To aid in continual learning, the memory also captures {\it mistakes} made by the agent in previous trials, similar to reflective agents explored in recent work \cite{reflexion, simulacra}, noting actions that failed to contribute to a task.

\textbf{Controller.}\hspace{0.7em} The role of the controller is to generate the next goal to pursue in service of the task.
In CLIN it is a frozen LLM, whose prompt includes the current {\bf task} $m$, e.g., ``convert water into steam'', retrievals from the {\bf current memory} $\mathcal{S}$, and the {\bf trial so far} (the sequence of goal-action-observation triples \{$g_1,a_1,o_1,...,g_t,a_t,o_t$\}), and is prompted to output the next
{\bf goal} $g_{t+1}$ to pursue, e.g., ``find water''. Memory items are retrieved using both the task instruction and the trial history. The controller first selects one or more memory items given the current state and if they are useful for the next action to progress in the task. After that, it appends the learning, if selected, in context to generate a goal, otherwise the goal is generated based on the trial history (see full prompt in \Cref{fig:prompt:clin-base}).

\textbf{Executor.}\hspace{0.7em} The role of the executor is to convert the generated goal $g_{t+1}$ into a valid {\bf action} $a_{t+1}$ that can be executed in the environment in pursuit of that goal.
Again a (frozen) LLM is used, whose prompt includes the goal $g_{t+1}$, the trial so far, and all the possible actions that can be performed in the
current state (provided by the simulator, as is standard practice in current generative agent research \cite{Ahn2022DoAI, Yao2022ReActSR, Lin2023SwiftSageAG, simulacra}).
The list of possible actions is expressed as possible action templates and available objects that can instantiate them, rather than
a combinatorially large enumeration of possible actions. The model is then prompted to generate a candidate action to perform (see prompt in \Cref{fig:prompt:clin-base}). Finally, CLIN checks this
candidate action is one of the valid actions. If it is not, it finds the most similar valid action using the pre-trained embeddings from the \texttt{sentence-transformer} model \cite{reimers-2019-sentence-bert}. If the top-ranked valid action has a similarity score greater than a threshold (here, 0.9, chosen as a hyperparameter), the action is selected. Otherwise, we perform iterative refinement \cite{Madaan2023SelfRefineIR} by suffixing the context with feedback that the generated candidate action is not executable. 
This allows the executor to retry the generation for  up to a maximum number of tries (here, 5).

\begin{figure}[t]
\begin{minipage}[h]{.5\textwidth}
\begin{algorithm}[H]
\begin{algorithmic}[1]
\small{
\Procedure{Adaptation}{Task: $m$, Env: $e$, Memory: $\mathcal{S}$}:
\State Initialize Memory: $\mathcal{S}_0$
\State {\bf for} $k \in 1, \cdots, K$ {\bf do}: 
\State \gapx Intialize Trial $\mathcal{T}$, $t$
\State \gapx {\bf while} $t <$ max. steps or task not complete {\bf do}:
\State \gapx \gapx $g_{t} =$ \texttt{{Controller}}~($m$, $e$, $\mathcal{T}_{<t}$, $\mathcal{S}_{k-1}$)
\State \gapx \gapx $a_{t} =$ \texttt{{Executor}}~($g_{t}$, admissible actions) 
\State \gapx \gapx $r_{t}, o_{t} =$ \texttt{{Simulator}}~($\mathcal{T}_{<t}$, $a_{t}$)
\State \gapx \gapx $\mathcal{T}_{<t+1} =$ $\mathcal{T}_{<t}$ + ($g_{t}$, $a_{t}$, $o_{t}$, $r_{t}$)
\State \gapx Final reward $r_{\text{k}}$ = $r_{t}$
\State \gapx $\mathcal{S}_{k} =$ \texttt{memory-generator}~($\{\mathcal{S}_{<k}\}$, $\mathcal{T}_k$, $r_k$)
\EndProcedure
\State

\Procedure{Generalization}{Task: $m$, Env: $e$, past $m'$/$e'$}
\State $\{\mathcal{S}_{\text{crucial}}, r_k\} =$  \texttt{crucial-memories}~(past $m'$/$e'$)
\State $\mathcal{S}_\text{meta} =$  \texttt{meta-memory} ($\{\mathcal{S}_{\text{crucial}}, r_k\}$, $m$)
\State \textsc{Adaptation}($m$, $e$, $\mathcal{S}_\text{meta}$)
\EndProcedure
}
\end{algorithmic}
\caption{Continual Learning with CLIN \label{clin-algo}}
\end{algorithm}
\end{minipage}%
\hspace{0.8em}
\begin{minipage}[h]{.48\textwidth}
 \centering
    \includegraphics[trim= 50 85 950 50,clip, width=\textwidth]{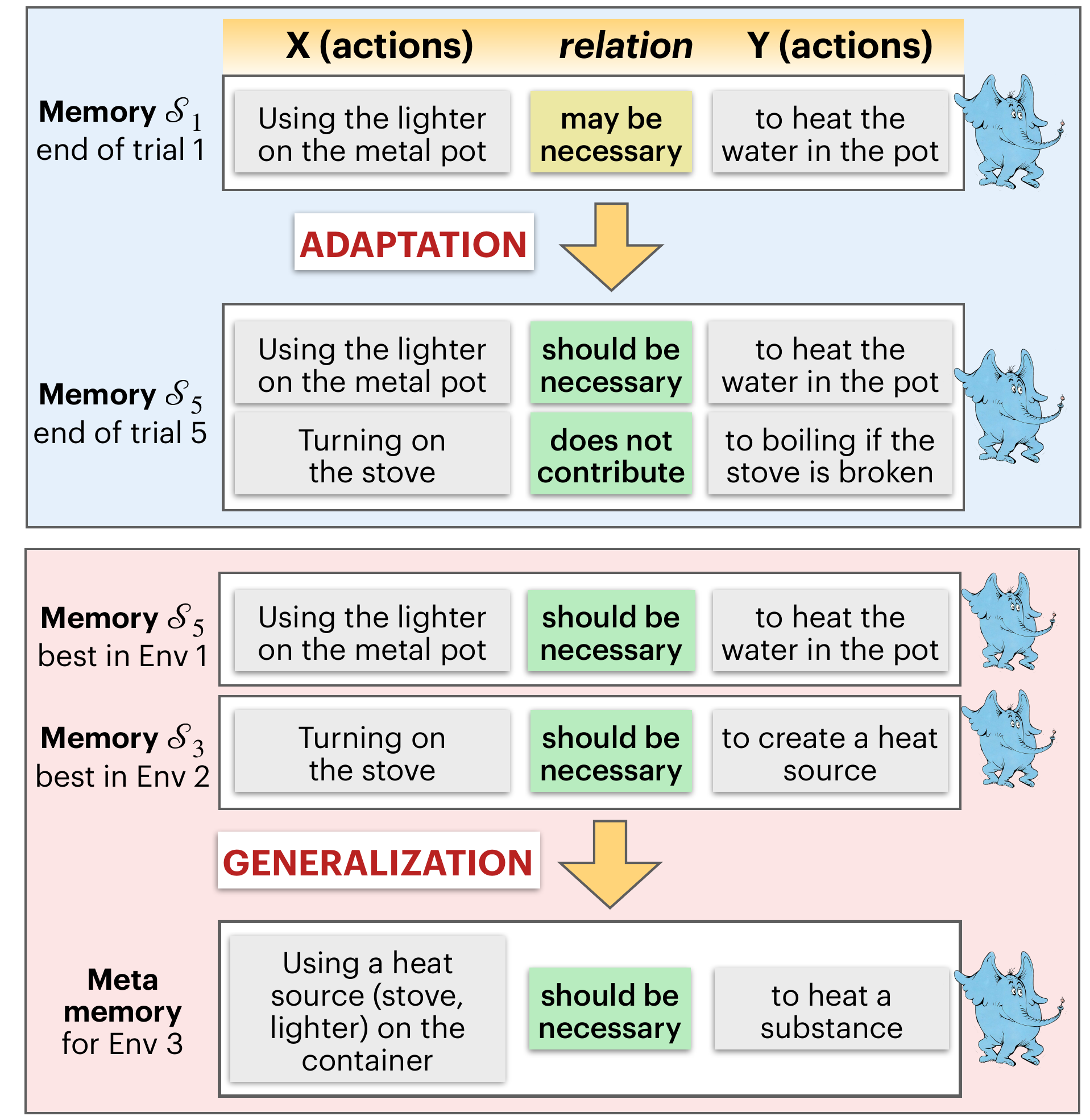} \\
\end{minipage}
 \caption{(\textsc{left}) \clin's continual learning algorithm. (\textsc{right}) Example causal abstractions. }
    \label{fig:algo-memory}
\end{figure}

Finally, upon executing the action $a_{t+1}$, \clin~receives a partial next state, as an {\bf observation}, from the simulator and the reward ($r$) $\in [0, 1]$. Rewards are nominally given by the simulator for achieving either major task subgoals (e.g., finding water, for the boiling task), or minor and optional subgoals (e.g., being in the kitchen, for the boiling task). Rewards are sparse and generally only supplied after the completion of a task subgoal. A snapshot of a full trial is given in lines 4-10 in Algorithm~\ref{clin-algo}.

Note that CLIN does not make use of any gold data to identify goals and memories. Rather, we expect \clin~to perform a balanced act of exploration-exploitation by interacting, learning, and adapting to unseen tasks or environment configurations---a key difference from few-shot generative agents by previous work \cite{Ahn2022DoAI, Yao2022ReActSR, Lin2023SwiftSageAG, simulacra}.

\subsection{Memory: A collection of Adaptive Causal Abstractions}
\label{sec:memory}

At the end of each trial (completion or failure), CLIN uses a {\bf memory generator} to create or update its memory.
The memory generator is a (frozen) LLM prompted to reflect on the current trial and memory, and generate a new memory
of insights in the form of (English sentences expressing) useful {\bf causal abstractions}, as we now describe.

Learning about state transitions is essential for sequential decision-making tasks \cite{Mnih2013PlayingAW}, which can be manifested by knowing 1) actions enabling desired state transitions, 2) actions producing undesired or no change in states, and 3) state transitions that contribute to the task progress. To generate these kinds of knowledge, the generator is prompted to generate insights in a particular syntax (see prompt in \Cref{fig:prompt:clin-adapt}). To capture good actions enabling desired changes and helpful state transitions, we use the template ``\texttt{X is NECESSARY to Y}'', and to capture contrastive examples of unsuitable actions and state transitions, we employ ``\texttt{X DOES NOT CONTRIBUTE to Y}'', as depicted in \Cref{fig:memory}, where \texttt{X}, \texttt{Y} are related to actions. These abstractions are functionally analogous to hindsight experience replay \cite{Andrychowicz2017HindsightER},
obtained from \clin's~past self-explorations.

While useful insights can be abstracted from the trials, \clin's exploration can be limited, especially in the early trials, given an incredibly large action space. Hence, incomplete exploration can pose varying degrees of uncertainty on extracted insights. To capture this, we also include a measure of uncertainty in each abstraction by either of the two linguistic variations in their surface forms: ``\texttt{X may ...}'' to denote moderate to high uncertainty, and ``\texttt{X should ...}'' to indicate low uncertainty (See \Cref{fig:memory}). In the course of continual learning, as \clin~gathers experience, we expect the level of uncertainty to change depending on the frequency of their use and their fitness to the task.

\textbf{Updating Memory Across Trials.} \clin~continually attempts to solve a task in an environment for multiple trials (in sum, an episode). To update the memory after each trial within an episode, the memory generator is prompted with the most recent trial (a sequence of ($g_t$, $a_t$, $o_t$) tuples and the final reward $r_k$\footnote{
The reward is converted to NL feedback for a LLM using 7 simple rules, e.g., ``if score >= 0 and score < 20 then feedback = {\it "The agent performed poorly and made some progress but not enough to solve the task."}}
),
and the memories from the three most recent trials $\{\mathcal{S}_{k-2},\mathcal{S}_{k-1},\mathcal{S}_{k}\}$. It is then prompted to generate an updated memory $\mathcal{S}_{k+1}$, namely a new list of semi-structured causal abstractions in the forms described above, for use in the next trial. Although we do not specify a maximum size for the memory, we observe that size of the generated memory (i.e., the number of causal abstractions generated) is far less than the number of actions executed in the trial, indicating the memory-generator additionally performs a saliency-based pruning to keep only important insights based on the success of the trial, as indicated by the final reward $r_{k}$ at the end of the trial $\mathcal{T}_k$.

\subsection{Meta-Memory for Better Generalization}
\label{sec:metamemory}
Updating memory based on past insights and the current trial to influence future trials for the same task in the current environment configuration during test-time adaptation. However, to generalize across tasks or environment configurations, the memory needs to contain more generalized causal abstractions than memories used across trials in an episode. We call this as \textbf{meta-memory}, abstracted across multiple episodes of solving different tasks in different environment configurations to be applicable in future episodes. 

\textbf{Auto-curriculum Selection.} Before we generate the meta-memory, it is important to choose memories extracted from the best trials from previous episodes because random sampling may not benefit \clin~in zero-shot generalization \cite{Team2023HumanTimescaleAI}. Following the prioritized level replay scheme \cite{Jiang2021ReplayGuidedAE}, we choose the most successful trial per episode and retrieve memories abstracted from those trials with a fixed archive of size 10, a hyperparameter.

\textbf{Generating Meta-Memory.} The goal of the meta-memory is to help \clin~generalize to unseen tasks and/or environments. While we keep the format of the causal abstractions the same as memories generated across trials, the prompt to generate the meta-memory is different than those used for generating per-trial memory. When the new memory is to be used for the {\it same} task but in a {\it different} environment, the prompt instructs for a meta-memory helpful ``\texttt{to solve the same task in a new environment configuration}'' given the target task description with an expectation that meta-memory abstractions will entail generic causal insights about the task irrespective of environment configurations (see \Cref{fig:prompt:clin-gen-env}). Similarly, when the new memory is to be used for a {\it different} task, the prompt is modified accordingly to reflect this (\Cref{fig:prompt:clin-gen-task}). Along with the target task description for better memory generation, each past memory selected to generate the meta-memory is attached to the final rewards for the associated trials, allowing the generator to combine insights across episodes and assign the levels of uncertainty using the evidence of success. 

\eat{
\textbf{Generating Meta-Memory.} The goal of the meta-memory is to aid \clin~generalize in unseen tasks or environment configurations. While we keep the format of the causal abstractions the same as memories generated across trials, the prompt to generate the meta-memory is different than those used for generating per-trial memory. Similar to \Cref{sec:memory}, we use an LLM-based generative module, now with different generalized instructions to generate the meta-memory. For example, if our generalization setup requires \clin~to generalize in a new environment configuration for a task attempted previously, the instruction explicitly mentions generating a meta-memory that can help ``\texttt{the agent to solve the same task in a new environment configuration}'' with an expectation that meta-memory abstractions will entail generic causal insights about the task irrespective of environment configurations. For each memory selected to generate the meta-memory, we also attach the final rewards for the associated trials, allowing the generator to combine insights across episodes and assign the levels of uncertainty depending on the evidence of success for the previous trials. 
}

\section{Results and Analysis}\label{experiment-setup}
\textbf{Experimental Setup.}\hspace{0.7em}
Test-time adaptation and generalization via continual learning require a variety of complex tasks and environment configurations to allow an agent to explore, learn latent causal insights from interactions, and exploit them in the future. We choose ScienceWorld \cite{Wang2022ScienceWorldIY}, a text-based interactive environment requiring complex interactive reasoning processes to solve a plethora of science-theory-based tasks spanning several diverse classes (e.g., thermodynamics, genetics, friction, etc.). The virtual space presents 10 sub-places: 
foundry, greenhouse, outside area, an art studio, workshop, kitchen, living room, bedroom, bathroom, and a hallway connecting inside rooms. The presence of several objects, their individual states, and action templates renders the search space intractable for any agent. ScienceWorld presents strikingly different environment configurations across task types, making it a rich testbed for evaluating adaptation and generalization. ScienceWorld tasks are partitioned into Short (S), e.g., \textit{pick \& place} and Long (L), e.g., \textit{grow plant}, tasks based on the number of required actions to succeed. 

Here, we define our setups for zero-shot adaptation (\textsc{Adapt}) and generalization (\textsc{{Gen-Env}} and \textsc{{Gen-Task}}). For all setups, we test \clin~and competing baselines on 18 tasks (two task instances from 9 classes) in several environment configurations from the test split of the ScienceWorld benchmark resulting in a total of 164 task-environment combinations unless stated otherwise. We evaluate based on the final reward score provided by the ScienceWorld simulator. 

\textsc{\textbf{Adapt}}: This setup focuses on \clin's ability to adapt to a task by attempting it for several trials in the same environment configuration. Most importantly, \clin~initializes with an empty memory at the beginning of the first trial and generates memory at the end of each trial. While the environment gets reset at the trial boundary, \clin's memory continues to be updated, capturing informative causal abstractions pertaining to both successful and failed actions. Here, we compare with Reflexion \cite{reflexion}, a SOTA, however, \clin~differs from Reflexion by how the memory is abstracted. 

\textsc{\textbf{Gen-Env}}: In this setup, we focus on \clin's ability to transfer its learning from past experiences to solve tasks in an unseen environment. For a task $m$, we run \clin~for 10 different (train) environment settings (with varying objects and starting locations) and then create meta-memories from its exploration to solve the same task in an unseen (test) environment. Here, we compare \clin~with RL methods 
DRRN \cite{He2015DeepRL}, KG-A2C \cite{Ammanabrolu2020GraphCR}, and CALM \cite{Yao2020KeepCA} trained on all (large) training variations with simulator reward and Generative Language agents, SayCan \cite{Ahn2022DoAI}, ReAct \cite{Yao2022ReActSR}, 
and Reflexion \cite{reflexion}, prompted with few-shot demonstrations.

\textsc{\textbf{Gen-Task}}: In this setup, we focus on \clin's ability to transfer its learning from past experiences to solve a new task in the same environment. For an environment $e$, we run \clin~for to solve a task $m$ and then condense its learning to solve a novel task $m'$ in the environment $e$. We took all test examples where we have a different task defined in the same environment configuration. \cite{Team2023HumanTimescaleAI} suggests that transferring learning from a random task can be very hard; hence we couple tasks that are related (revolve around overlapping task-critical objects/locations such water, kitchen), such as {\it boil} and {\it freeze} to measure transfer learning from one to the other. This is a novel setup where we do not have any off-the-shelf baselines. However, here, we compare against \clin-\textsc{Base}, a strong baseline agent. 

\textbf{\genadapt} (\textsc{G+A}): If \clin{}, in \genenv{} or \gentask{} setting, does not successfully complete the new task, it can continue learning and retrying that task. We refer to this setup as \genadapt.  
\clin{} can use any instruction-tuned LLM \cite{Chung2022ScalingIL} as part of the controller, executor, and memory generator. In this paper, we use \texttt{gpt-4}, the same as our generative agent baselines.

\begin{figure}[t!]
\begin{minipage}[h]{.7\textwidth}
\centering
\includegraphics[trim= 310 290 320 250,clip, width=\textwidth]{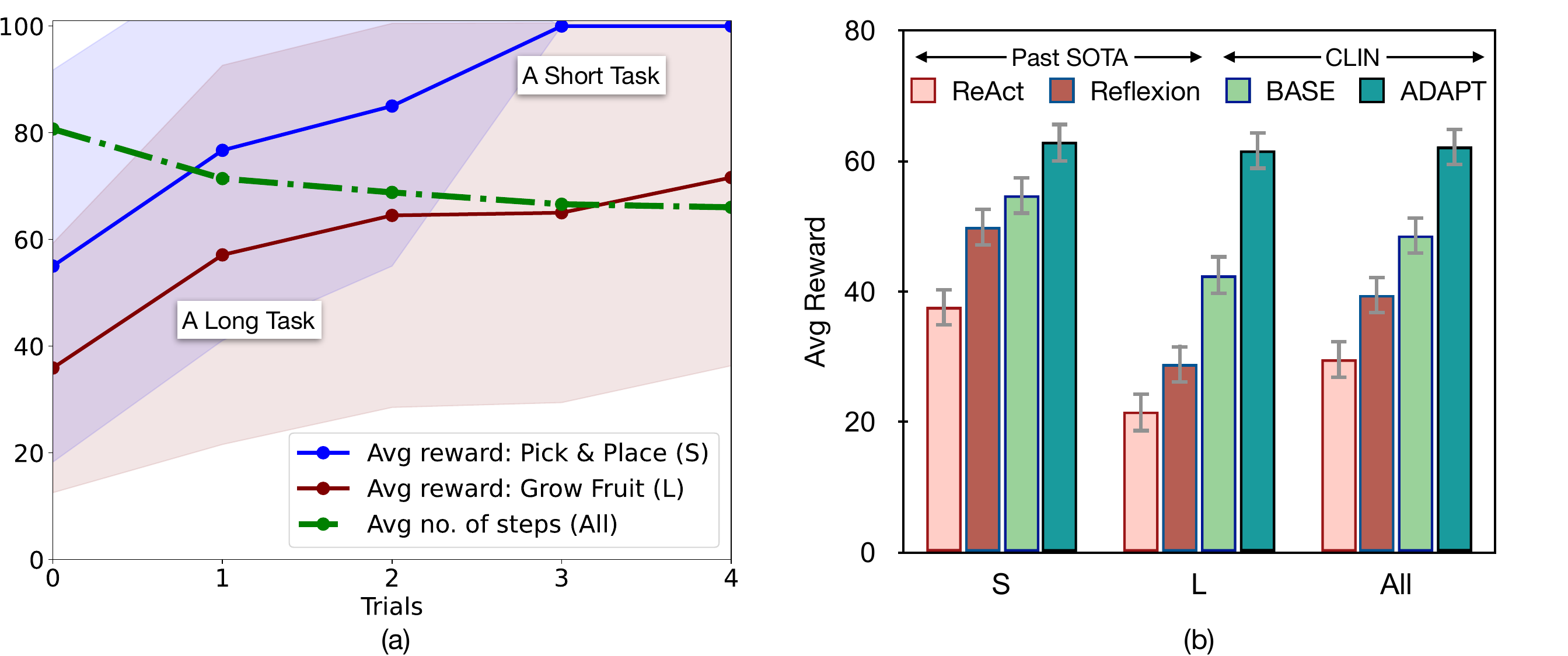} 
\end{minipage}%
\begin{minipage}[h]{.3\textwidth}
\centering
{\small
\resizebox{\linewidth}{!}{%
\begin{tabular}{@{}lcc@{}}
\multicolumn{3}{c}{\bf \clin's \textsc{Adapt} improvements}\\
\toprule
\bf Type & \bf \#trials to & \bf \%ep.  \\
\bf  & \bf  success ($\downarrow$) & \bf improv. \\
\midrule
S	& 3.3 & 29.2	 \\
L	& 3.2 & 37.2 \\
All & 3.3 & 33.2 \\
\bottomrule
\end{tabular}%
} 
\\
(c) 
}
\end{minipage}
\vspace{-3mm}
\caption{\textbf{Rapid task adaptation with \clin.}
\textbf{(a)} Example tasks where \clin{} improves scores across trials. For \clin, Trial-0 is the \textsc{Base}, Trial-4 is the \textsc{Adapt}. \textbf{(b)} Comparison of \clin{} with Reflexion \cite{reflexion}.
\textbf{(c)} \clin~improves from \textsc{Base} to \textsc{Adapt} (full results in \Cref{sec:more_results}). \label{q1-adapt-combined}}
\end{figure}

\subsection{\clin~Exhibits Rapid Task Adaptation}

\Cref{q1-adapt-combined}a demonstrates two example trends where \clin{} learns from its own prior attempts (\textsc{Adapt}) and gets better at solving a given task. Apart from length, the difficulty level of a task also depends on the environment configuration (hence, variance across environment configurations for each task). \clin{} quickly adapts to a short task, \textit{Pick \& Place}, solving it in its 4th attempt, whereas for a longer task, \textit{Grow Fruit}, it is not solved after 5th (max) attempts.
Furthermore, \Cref{q1-adapt-combined}a depicts, \clin~becomes more efficient in later trials by solving the tasks with a lower number of (average) steps. Figure \ref{q1-adapt-combined}c shows an average number of attempts\footnote{During \textsc{Adapt}, \clin~tries up to 5 trials. If it solves a task with a score of 100, it stops retrying.} for \clin~to solve a task and \% episodes per task where scores improved compared to its own first trial. 

Next, we compare \clin{} with Reflexion, the reflective SOTA agent, in \Cref{q1-adapt-combined}b. \clin~already starts off with a stronger base performance (see discussion in \ref{sect:discussions}), however, \clin's~ relative improvement in \textsc{Adapt} is significantly stronger than Reflexion's gain from its base agent ReAct. Furthermore, \clin's relative improvement is higher for longer tasks. 
This can be attributed to \clin's persistent memory, which gets refined over past trials, whereas Reflexion may fall short of collecting useful learnings from earlier trials as it only focuses on the current trial for its reflections (hence not long-term). Furthermore, \clin{} accumulates both useful (for the task) and harmful (for the task) causal learnings, whereas Reflexion only learns from its mistakes, lacking comprehensive learning. 



\begin{table}[t!]
\centering
{\small
\resizebox{0.9\linewidth}{!}{%
\begin{tabular}{@{}l@{}c||ccc|ccc|cc|c@{}}
\toprule
  & & \multicolumn{3}{c|}{\bf RL Methods} & \multicolumn{3}{c|}{\bf Generative Language Agents} & \multicolumn{3}{c}{\bf \clin{ (ours)}} \\
 \midrule
\bf Task &  \bf Type & \bf DRRN & \bf KGA2C & \bf CALM & \bf SayCan & \bf ReAct & \bf Reflexion & \bf \textsc{Base} & \bf \textsc{Gen-Env} & \bf \textsc{G+A} \\\midrule
 Temp\textsubscript{1} & S & 6.6 & 6.0 & 1.0 & \textbf{26.4} & 7.2 & 5.9 & 25.2 & 15.7 & 13.8 \\
 Temp\textsubscript{2} & S & 5.5 & 11.0 & 1.0 & 8.0 & 6.1 & 28.6 & 53.2 & 49.7 & \textbf{58.2} \\
 Pick\&Place\textsubscript{1} & S & 15.0 & 18.0 & 10.0 & 22.9 & 26.7 & 64.9 & 92.5 & 59.2 & \textbf{100.0} \\
 Pick\&Place\textsubscript{2} & S & 21.7 & 16.0 & 10.0 & 20.9 & 53.3 & 16.4 & 55.0 & \textbf{100.0} & \textbf{100.0} \\
 Chemistry\textsubscript{1} & S & 15.8 & 17.0 & 3.0 & 47.8 & 51.0 & \textbf{70.4} & 44.5 & 42.2 & 51.7 \\
 Chemistry\textsubscript{2} & S & 26.7 & 19.0 & 6.0 & 39.3 & 58.9 & 70.7 & 56.7 & 85.6 & \textbf{93.3} \\
 Lifespan\textsubscript{1} & S & 50.0 & 43.0 & 6.0 & 80.0 & 60.0 & \textbf{100.0} & 85.0 & 65.0 & \textbf{100.0} \\
 Lifespan\textsubscript{2} & S & 50.0 & 32.0 & 10.0 & 67.5 & 67.5 & 84.4 & 70.0 & 75.0 & \textbf{90.0} \\
 Biology\textsubscript{1} & S & 8.0 & 10.0 & 0.0 & 16.0 & 8.0 & 8.0 & 10.0 & 32.0 & \textbf{32.0} \\
 Boil & L & 3.5 & 0.0 & 0.0 & \textbf{33.1} & 3.5 & 4.2 & 7.0 & 4.4 & 16.3 \\
 Freeze & L & 0.0 & 4.0 & 0.0 & 3.9 & 7.8 & 7.8 & \textbf{10.0} & 8.9 & \textbf{10.0} \\
 GrowPlant & L & 8.0 & 6.0 & 2.0 & 9.9 & 9.1 & 7.3 & 10.2 & 10.9 & \textbf{11.2} \\
 GrowFruit & L & 14.3 & 11.0 & 4.0 & 13.9 & 18.6 & 13.0 & 35.9 & 70.8 & \textbf{94.5} \\
 Biology\textsubscript{2} & L & 21.0 & 5.0 & 4.0 & 20.9 & 27.7 & 2.6 & 70.0 & 42.8 & \textbf{85.6} \\
 Force & L & 10.0 & 4.0 & 0.0 & 21.9 & 40.5 & 50.6 & 53.5 & 70.0 & \textbf{100.0} \\
 Friction & L & 10.0 & 4.0 & 3.0 & 32.3 & 44.0 & \textbf{100.0} & 56.5 & 70.0 & 94.0 \\
 Genetics\textsubscript{1} & L & 16.8 & 11.0 & 2.0 & 67.5 & 25.7 & 50.9 & 77.4 & 84.5 & \textbf{100.0} \\
 Genetics\textsubscript{2} & L & 17.0 & 11.0 & 2.0 & 59.5 & 16.8 & 23.7 & 62.3 & 61.4 & \textbf{100.0} \\
\midrule
 & \textbf{S} & 22.1 & 19.1 & 5.2 & 36.5 & 37.6 & 49.9 & 54.7 & 58.3 & \textbf{71.0} \\
 & \textbf{L} & 11.2 & 6.2 & 1.9 & 29.2 & 21.5 & 28.9 & 42.5 & 47.1 & \textbf{68.0 }\\
 & \textbf{All} & 16.7 & 12.7 & 3.6 & 32.9 & 29.6 & 39.4 & 48.6 & 52.7 & \textbf{69.5} \\
\bottomrule
\end{tabular}%
}
}
\caption{Comparing \clin{} with baselines for \textbf{generalization across unseen environments} \label{table:q2-generalization}}
\end{table}

\subsection{\clin{} Outperforms SOTA, Generalizing to Novel Environments and Tasks}
\textbf{Generalizing to new environments.}\hspace{0.7em} \Cref{table:q2-generalization} compares \clin{} with baselines that learn from training environmental variants for a task to improve its performance in a novel environment \footnote{Baseline numbers are derived from 
Table 1 in \cite{Lin2023SwiftSageAG}}. Language agents (including \clin) that use NL feedback from the ScienceWorld (e.g., ``Door to the kitchen is closed'') perform significantly better compared to RL methods that purely rely on (sparse) numeric rewards from the environment to learn a policy. 
We observe a positive generalization effect in \genenv~(average 4 point gain) compared to \textsc{Base} where \clin~tries to solve the tasks zero-shot. With a strong \textsc{Base} performance, \clin~beats all baselines in generalization performance. Furthermore, in \textsc{G+A}, \clin~shows a substantial 16 additional improvement, beating the SOTA reflective agent by 23 points. 
Figure \ref{fig:gen-task}a additionally shows trend of improvement compared to when \clin~does not start with a meta-memory. Meta-memory helps \clin{} with a stronger start than \textsc{Base} (52.7 vs. 48.6), with a continued gain in scores till the end of Trial-4 (\textsc{G+A}: 69.5 vs. \textsc{Adapt}: 62.2). The stronger start for \clin~with meta-memory also results in fewer steps to solve a task.
Unlike imitation learning-based agents, TDT \cite{Wang2022ScienceWorldIY} and SwiftSage \cite{Lin2023SwiftSageAG}, \clin{} (and most baselines) does not use any gold trajectories. Refining its memory only from self-generated trajectories, \clin~outperforms TDT on all 18 tasks and SwiftSage on 8/18 (mostly long) tasks.  


\textbf{Generalizing to new tasks.}\hspace{0.7em}
Mirroring trends from \genenv, \clin~ demonstrates strong transfer learning to new tasks with 13-point improvement over its \textsc{Base} performance, being better at 38.8\% of times (\Cref{fig:gen-task}c). The improvement attributes to critical learning about the environment (``apple juice is in the fridge'', required for both boiling and freezing it), leading to improvement in previously low-performing tasks in both \textsc{Adapt} and \textsc{Gen-Env} setups. This transfer learning in \gentask{} and \textsc{G+A} helps \clin{} to solve the tasks with a lesser number of steps\footnote{\# steps in \Cref{fig:gen-task}a,b are normalized between 0-1, 1 being maximum \# steps allowed for a task.} and achieve higher rewards.

\subsection{Discussion} \label{sect:discussions}
\begin{figure}[t!]
\begin{minipage}[h]{.5\textwidth}
\centering
    \includegraphics[trim= 50 115 1000 55,clip, width=0.90\textwidth]{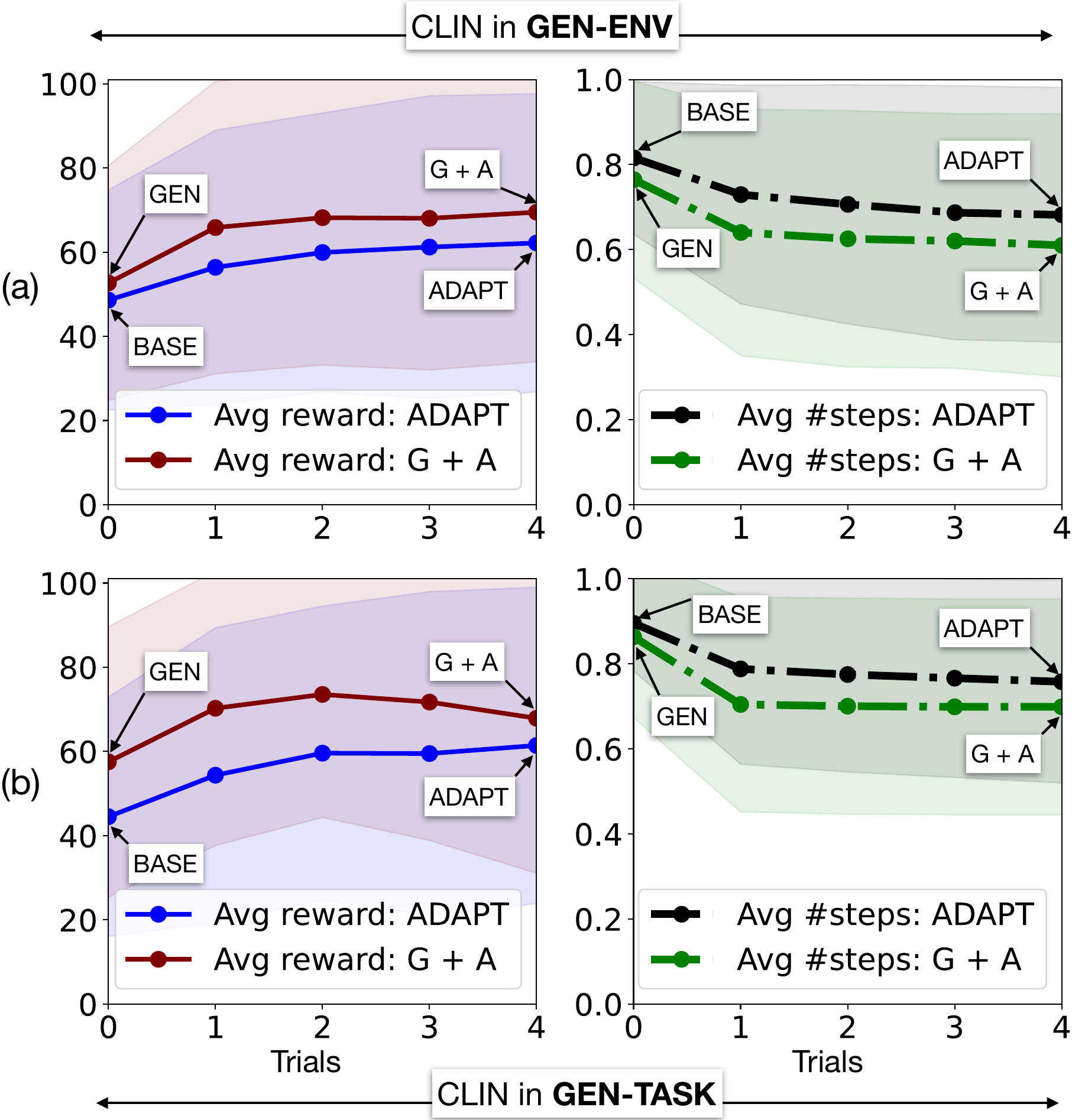}
    \label{fig:memory}
\end{minipage}%
\begin{minipage}[h]{.5\textwidth}
\centering
{\small
\resizebox{0.8\linewidth}{!}{%
\begin{tabular}{@{}l||cc|cc@{}}
\multicolumn{5}{c}{\bf \clin's \textsc{Gen-Task} improvements}\\ \toprule
\bf Type &   \multicolumn{2}{c|}{\bf \textsc{Gen-Task}} &  \multicolumn{2}{c}{\bf \textsc{G + A}} \\
& \bf $\Delta$avg & \bf \%ep. & \bf $\Delta$avg & \bf \%ep. \\
& \bf score & \bf improv. & \bf score & \bf improv. \\
\midrule
S &  14.6 & 40.0  &  4.9 & 5.7  \\
L &  10.3 & 36.7  & 9.2 & 15.6  \\ \midrule
All & 13.0 & 38.8 &  6.5 & 9.3  \\
\bottomrule
\end{tabular}%
}
}
\\
(c) 
\\
\vspace{1.5em}


\begin{minipage}[h]{\textwidth}
\centering
{\small
\resizebox{0.9\linewidth}{!}{%
\begin{tabular}{@{}l||cc@{}}
\multicolumn{3}{c}{\bf Ablations for \clin}\\ 
\toprule
\bf Ablation Setup & \bf $\Delta$avg & \bf \%ep.\\
&  \bf score ($\downarrow$) & \bf drop. ($\uparrow$)\\
\midrule
Abl-Causal-Memory & -6.2 & 10 \\
Abl-Controller-\textsc{Base} &  -18.1 & 44.8 \\
\bottomrule
\end{tabular}%
}
}
\\
(d)

\end{minipage}
\end{minipage}

\caption{Reward and \#steps trends for \clin{} in \textbf{(a)} \genenv{} and \textbf{(b)} \gentask. 
\textbf{(c)} \% episode improvements and score change than \clin~without meta-memory (\gentask). \textbf{(d)} \clin~ablations.
}
\label{fig:gen-task}
\end{figure}

\textbf{Importance of memory structure. }\hspace{0.5em} 
\clin{}~extracts causal abstractions structured around `necessary' and `does not contribute' relations. As an ablation study, we modified our memory generator to generate free-form advice for future trials (without any constraint on their formats). 
We find that the average reward drops by 6 points (in 10\% cases compared to \clin) when using the unstructured memory, indicating the usefulness of causal abstractions, as shown in \Cref{fig:gen-task}d.

\textbf{Superior \textsc{Base} performance.}\hspace{0.5em} \Cref{q1-adapt-combined} depicts a superior \textsc{Base} performance for \clin~than the final performance of both ReAct and Reflexion despite using the same underlying LLM (here, \texttt{gpt-4}). We find if we ablate for the controller module in \clin, responsible for generating a goal before outputting the next action, \clin's \textsc{Base} performance drops in 44\% cases. With an 18 point drop in average reward (see \Cref{fig:gen-task}d), Abl-Contoller-\textsc{Base} version of \clin~becomes equivalent to ReAct, the base agent for Reflexion, demonstrating the importance of controller even in \textsc{Base} setup.


\textbf{A qualitative example.}\hspace{0.7em} 
\Cref{fig:algo-memory} depicts how memory items get refined during task adaptation and for generalization for a task \textit{boil}. \texttt{Env2} has a working stove, whereas in \texttt{Env1}, the stove is broken, but a lighter is available as an alternative. With a number of trials in these environments, \clin~learns how to use these two devices to generate heat. In an unseen environment with a broken stove, \clin~quickly receives a positive reward by using a lighter to heat a substance. While insights within an episode are often specific, e.g., ``Using the \textit{lighter} on the metal pot should be necessary to heat the \textit{water} in the pot'', \clin{} compiles these insights for a new target environment (as meta-memory), e.g., ``Using a \textit{heat source} (stove, lighter) on the container should be necessary to heat a \textit{substance}.'' \Cref{sec:appendix_mem} contains examples of memories generated during adaptation and generalization.

\textbf{Limitation: Lack of exploration.}\hspace{0.5em} \clin's learnings are dependent on its own past experience. If \clin~never explores a location in the environment or does not perform an action, an insight related to the unobserved activity can never be generated. Hence, exploration becomes important when task-critical location or action in unknown to \clin~from past trials. For example, in task of creating an orange  paint, the agent is supposed to find red and yellow paints from the art studio. However, art studio is not visible when \clin~starts from location `outside'. Unless the \clin~knows that there exists an art studio, it tries alternative method to create orange paints from other irrelevant objects (e.g., an orange) and remains unsuccessful. When a memory related art-studio appears from past exploration, \clin~is able to successfully complete the task. Similarly, in boil or freeze tasks, \clin~is unable to perform well which requires it to consistently measure the temperature of the substance to know its boiling/freezing point---an act it could never perform successfully in past trials resulting into less useful memory insights and subsequent lower performance in future trials.


\textbf{Limitation: Poor memory retrieval.}\hspace{0.5em} 
For a task of boiling gallium, \clin~is supposed to use oven/blast furnace and not a stove. In the meta-memory for boiling tasks, there are two insights regarding the act of boiling: ``Activating stove should be necessary to boil a substance'' and ``Using an alternative heat source (e.g., oven or fire pit) may be necessary if the initial heat source is insufficient.'' However, \clin~repeatedly retrieves the former and hence failing at the task despite performing other actions (e.g., finding gallium) correctly. This problem intensifies at the initial trial during generalization due to the presence of insights with varied initial conditions for them to be applied. 
This can be circumvented by improved memory representation, which we leave as a future work.


\section{Conclusion}
Our goal is a system that can continually improve over time, both while rapidly adapting to a task by multiple retries and efficiently generalizing to novel tasks and environments.
We propose \clin, an architecture for language agents that constructs a persistent, dynamic memory of causal abstractions, refines it over time and uses it effectively to improve its performance on future tasks, achieving state-of-the-art performance. Our work systematically evaluates a novel nonparametric learning paradigm, promising never-ending learning abilities to frozen language agents.  

\paragraph{Acknowledgement}
We sincerely thank Aristo team members Tushar Khot, Ashish Sabharwal, Shashank Gupta, Nathaniel Weir, Kyle Richardson, Jiangjie Chen, Archiki Prasad, and other members such as Faeze Brahman, Alexander Koller at the Allen Institute of AI for their generous feedback.

\bibliography{references}
\bibliographystyle{iclr2024_conference}

\clearpage
\newpage
\appendix

\section{\clin~prompts}
\label{appendix:clin-prompts}
\Cref{fig:prompt:clin-base,fig:prompt:clin-adapt,fig:prompt:clin-gen-env,fig:prompt:clin-gen-task} are the complete prompts for next-action generation (controller + executor), memory-generator during \textsc{Adapt}, \genenv, and \gentask.

\begin{figure}[h]
\begin{small}
    \begin{minted}[fontsize=\footnotesize, frame=lines, framesep=2mm, baselinestretch=1.2, breaklines, breaksymbolleft={}, breaksymbolright={},bgcolor=Box1Color]{text}
[System]: You are an AI agent helping execute a science experiment in a simulated environment with limited number of objects and actions available at each step.

[User]: 
Possible objects ( value an OBJ can take ): 
{objects_str}

Your next action should be in one of the following formats:
Possible actions:
{actions_str}

If I say \"Ambiguous request\", your action might mean multiple things. In that case, respond with the number corresponding to the action you want to take.

What action would you like to do next?

First, scan the (unordered) list of learnings, if provided. Decide if any of the learnings are applicable given the last observation to make progress in this task. Then only use selected learnings, if any, to construct a rationale for picking the next action. If no Learning is selected, construct the rationale based on the last observation. Format your response as follows:

Write 'I used learning id(s):' as a comma separated list; the list can be empty if no learnings selected. Then, write $$$ followed by the rationale. Finally, write ### followed by the single next action you would like to take.

If you think you have completed the task, please write TASK_COMPLETE as the next action.

If the task requires you to 'focus' on something (OBJ), please write FOCUS ON <OBJ> as the next action. FOCUS is a extremely critical action that can be only used the number of times 'focus' is mentioned in the task description. Using it more than that or inappropiately (such as on a wrong object) will terminate the session and the task will be rendered as incomplete.

If you performed an action that requires waiting to see the effect, please write 'wait' as the next action.
\end{minted}

\end{small}
\caption{Prompt for the Controller and the Executor}
    \label{fig:prompt:clin-base}
\end{figure}

\begin{figure}
\begin{small}
    \begin{minted}[fontsize=\footnotesize, frame=lines, framesep=2mm, baselinestretch=1.2, breaklines, breaksymbolleft={}, breaksymbolright={},bgcolor=Box2Color]{text}
[System]: You are an expert assistant.

[User]: 
You are given CURRENT TRACE, a sequence of actions that an agent made in a world to accomplish a task.

Task is detailed at the beginning.
For each action, there is a rationale why the agent made that action.
There is an observation that provide details about the new state of the world after each action was executed.
The CURRENT TRACE is accompanied by an EVALUATION REPORT indicating the success of the attempt to the task.

You can also be provided with PREVIOUS LEARNINGS which are learnings from the previous attempts by the agent for the same task in the same environment/world. TASK indicates the task description. EPISODE indicates the number of previous attempts of the task.

Generate a summary of learning, as a numbered list, that will help the agent to successfully accomplish the SAME task AGAIN, in the SAME world.

Each numbered item in the summary can ONLY be of the form:
X MAY BE NECCESSARY to Y.
X SHOULD BE NECCESSARY to Y.
X MAY BE CONTRIBUTE to Y.
X DOES NOT CONTRIBUTE to Y.

{CURRENT TRACE}
Action: ...
Observation: ...
...
EVALUATION REPORT:
REWARD_FINAL: 100. This means: The agent has performed exceptionally well and successfully solved the task.

Summary of learning as a numbered list:
\end{minted}

\end{small}
\caption{Prompt for \clin's memory generator during \textsc{Adapt}}
    \label{fig:prompt:clin-adapt}
\end{figure}

\begin{figure}
\begin{small}
    \begin{minted}[fontsize=\footnotesize, frame=lines, framesep=2mm, baselinestretch=1.2, breaklines, breaksymbolleft={}, breaksymbolright={},bgcolor=Box3Color]{text}
[System]: You are an expert assistant.

[User]: You are given a collection of learning lists, that are derived from actions made by an agent and subsequent observations from a world to accomplish a TYPE of TASKs. All of these TASKs belong to a same TYPE (such as 'boiling') but they are executed in different ENVIRONMENT configurations. A different ENVIRONMENT configuration means there are presence of a different set of objects (lighter instead of a stove) that are critical for solving the TASK, presence of a different set of distractor objects that are not useful for the TASK, a different floor plan, etc.

For each learning list, the TASK description is provided at the beginning as TASK:

Each learning list indicates a list of learnings from the agent's best attempt to solve the TASK.

Each learning list is associated with an EVALUATION REPORT indicated how sucessful the respective attempt was for solving the task.

Consider all learning lists and combine them in to a summary of learnings, as a numbered list, that will help the agent to successfully accomplish a NEW TASK related to the previous TASKs (such as 'boliing') in an ENVIRONMENT configuration that it has not seen before. The NEW TASK description will be provided.

Each numbered item in the summary can ONLY be of the form:
X MAY BE NECCESSARY to Y.
X SHOULD BE NECCESSARY to Y.
X MAY NOT CONTRIBUTE to Y.
X DOES NOT CONTRIBUTE to Y.

{PREVIOUS LEARNINGS}
TASK: ...
LEARNINGS:...
EVALUATION REPORT:
REWARD_FINAL: 100. This means: The agent has performed exceptionally well and successfully solved the task.
...

NEW TASK: ...
Summary of learning as a numbered list:
\end{minted}

\end{small}
\caption{Prompt for \clin's memory generator during \textsc{Gen-Env}}
    \label{fig:prompt:clin-gen-env}
\end{figure}

\begin{figure}
\begin{small}
    \begin{minted}[fontsize=\footnotesize, frame=lines, framesep=2mm, baselinestretch=1.2, breaklines, breaksymbolleft={}, breaksymbolright={},bgcolor=Box3Color]{text}
[System]: You are an expert assistant.

[User]: You may be given a list of learnings, that are derived from actions made by an agent and subsequent observations from a world to accomplish a TASK in an ENVIRONMENT CONFIGURATION.

For the learning list, the TASK description is provided at the beginning as TASK:

The learnings are from the agent's best attempt to solve the TASK.

The learning list is associated with an EVALUATION REPORT indicated how sucessful the attempt was for solving the task.

Now, generate a summary of learnings from the existing ones if provided, such that they will be useful to the NEW TASK in the SAME ENVIRONMENT CONFIGURATION. The NEW TASK may require different actions which are not captured in the given learnings but given learnings can be used to infer about the ENVIRONMENT CONFIGURATION. The NEW TASK description will be given. If PREVIOUS LEARNINGS says 'No learnings available', improvise learnings for the NEW TASK.

Each numbered item in the summary can ONLY be of the form:
X MAY BE NECCESSARY to Y.
X SHOULD BE NECCESSARY to Y.
X MAY NOT CONTRIBUTE to Y.
X DOES NOT CONTRIBUTE to Y.

{PREVIOUS LEARNINGS}
TASK: ...
LEARNINGS:...
EVALUATION REPORT:
REWARD_FINAL: 100. This means: The agent has performed exceptionally well and successfully solved the task.
...

NEW TASK: ...
Summary of learning as a numbered list:
\end{minted}

\end{small}
\caption{Prompt for \clin's memory generator during \textsc{Gen-Task}}
    \label{fig:prompt:clin-gen-task}
\end{figure}

\section{Example Memories}
\label{sec:appendix_mem}
Example generated memory for \textsc{Adapt}, \genenv, and \gentask setups in \Cref{fig:memory:clin-adapt,fig:memory:clin-gen-env,fig:memory:gen-task}.

\begin{figure}
\begin{small}
    \begin{minted}[fontsize=\footnotesize, frame=lines, framesep=2mm, baselinestretch=1.2, breaklines, breaksymbolleft={}, breaksymbolright={},bgcolor=Box2Color]{text}

Task: Your task is to determine whether round seed shape is a dominant or recessive trait in the pea plant. If the trait is dominant, focus on the blue box. If the trait is recessive, focus on the orange box.

Memory (ADAPT), after Trial5:
1. Looking around MAY BE NECESSARY to gather information about the environment.
2. Moving to the greenhouse SHOULD BE NECESSARY to find the pea seeds.
3. Picking the jar SHOULD BE NECESSARY to access the pea seeds.
4. Moving the round green pea seed to flower pot 2 SHOULD BE NECESSARY to grow the pea plants.
5. Moving the wrinkly green pea seed to flower pot 3 SHOULD BE NECESSARY to grow the pea plants.
6. Waiting SHOULD BE NECESSARY for the pea plants to grow and show their traits.
7. Looking at the adult pea plant in flower pots 2 and 3 MAY BE NECESSARY to observe the traits of the offspring.
8. Focusing on the blue box SHOULD BE NECESSARY to indicate the trait of the seed shape.
9. Continuously looking at the same environment DOES NOT CONTRIBUTE to the task of determining the dominant or recessive trait in the pea plant.
10. Moving to the workshop DOES NOT CONTRIBUTE to determining the dominant or recessive trait in the pea plant.
11. Picking up the battery or the wires in the workshop DOES NOT CONTRIBUTE to finding the pea seeds.
12. Waiting for longer periods MAY BE NECESSARY to allow the pea plants to fully grow and show their traits.
\end{minted}

\end{small}
\caption{Memory generated after Trial5 in \textsc{Adapt} for a Genetics task.}
    \label{fig:memory:clin-adapt}
\end{figure}

\begin{figure}
\begin{small}
    \begin{minted}[fontsize=\footnotesize, frame=lines, framesep=2mm, baselinestretch=1.2, breaklines, breaksymbolleft={}, breaksymbolright={},bgcolor=Box3Color]{text}

Task: Your task is to determine which of the two inclined planes (aluminum, platinum) has the most friction. After completing your experiment, focus on the inclined plane with the most friction.

Meta-memory (GEN-ENV):
1. Moving to the hallway SHOULD BE NECESSARY to reach the workshop.
2. Moving to the workshop SHOULD BE NECESSARY to find the block.
3. Picking up the block SHOULD BE NECESSARY to move it to the inclined planes.
4. Placing the block on the first inclined plane (either aluminum or platinum) SHOULD BE NECESSARY to measure the friction.
5. Activating the stopwatch SHOULD BE NECESSARY to time the experiment.
6. Waiting for a certain period MAY CONTRIBUTE to observing the friction effect.
7. Deactivating the stopwatch SHOULD BE NECESSARY to stop timing the experiment.
8. Moving the block to the second inclined plane (either aluminum or platinum) SHOULD BE NECESSARY to compare the friction.
9. Activating the stopwatch again SHOULD BE NECESSARY to time the second part of the experiment.
10. Waiting for a certain period again MAY BE NECESSARY to observe the friction effect.
11. Deactivating the stopwatch again SHOULD BE NECESSARY to stop timing the experiment.
12. Focusing on the inclined plane with the most friction SHOULD BE NECESSARY to conclude the experiment.
13. Repeating the experiment multiple times MAY BE NECESSARY for more accurate results.
14. Looking around in the initial room multiple times DOES NOT CONTRIBUTE to the task.
15. Moving the block back and forth between the two inclined planes DOES NOT CONTRIBUTE to the task.
\end{minted}

\end{small}
\caption{Meta-memory used in \genenv~for a Friction task.}
    \label{fig:memory:clin-gen-env}
\end{figure}

\begin{figure}
\begin{small}
    \begin{minted}[fontsize=\footnotesize, frame=lines, framesep=2mm, baselinestretch=1.2, breaklines, breaksymbolleft={}, breaksymbolright={},bgcolor=Box3Color]{text}

Task: Your task is to determine whether round seed shape is a dominant or recessive trait in the pea plant. If the trait is dominant, focus on the blue box. If the trait is recessive, focus on the orange box.

Meta-memory (GEN-TASK):
Task: Your task is to freeze mercury. First, focus on the substance. Then, take actions that will cause it to change its state of matter.

Meta-memory (GEN-TASK):
1. Looking around MAY BE NECESSARY to identify the available resources and the layout of the environment.
2. Moving to different rooms SHOULD BE NECESSARY to find the tools and materials needed to change the state of the substance.
3. Picking up items like glass cups or metal pots SHOULD BE NECESSARY to contain the substance for changing its state.
4. Focusing on the substance SHOULD BE NECESSARY to understand its properties and how to interact with it.
5. Picking up the thermometer SHOULD BE NECESSARY to monitor the temperature of the substance.
6. Using the thermometer on the substance SHOULD BE NECESSARY to monitor the progress of the task.
7. Puring the substance into the container SHOULD BE NECESSARY to prepare it for cooling.
8. Moving the container to a cooling device SHOULD BE NECESSARY to cool the substance.
9. Waiting for a period of time after cooling the substance SHOULD BE NECESSARY to allow the substance to change state.
10. Repeatedly checking the temperature of the substance SHOULD BE NECESSARY to monitor the progress of the task.
11. Activating the stove DOES NOT CONTRIBUTE to the task as it does not progress the task.
12. Picking up unrelated items like a lighter DOES NOT CONTRIBUTE to the task as it does not progress the task.
13. Moving to unrelated rooms like the workshop DOES NOT CONTRIBUTE to the task as it does not progress the task.
14. Teleporting to the kitchen MAY BE NECESSARY for the task as it speeds up the process of moving between rooms.
15. Using the thermometer multiple times on the substance after it reaches freezing point DOES NOT CONTRIBUTE to the task as it does not progress the task.
\end{minted}

\end{small}
\caption{Meta-memory used in \gentask~for a Freeze task.}
    \label{fig:memory:gen-task}
\end{figure}


\section{More results}
\label{sec:more_results}
Full results for \clin~outperforming Reflexion is in \Cref{table:q1}. For ScienceWorld benchmark, we exclude electricity tasks since they deviate from standard electrical conventions, prohibiting us from fairly using LLM agents. We choose the first 10 test variants for each 18 tasks selected. The full list of 18 tasks from the benchmark, with the number of test variants used in parentheses:

grow-plant (10), identify-life-stages-1 (5), grow-fruit (10), measure-melting-point-known-substance (10), mendelian-genetics-unknown-plant (10), chemistry-mix-paint-secondary-color (9), freeze (9),
lifespan-longest-lived (10), inclined-plane-determine-angle (10), boil (9), use-thermometer (10), chemistry-mix (8), lifespan-shortest-lived (10), find-plant (10), find-living-thing (10), identify-life-stages-2 (4), mendelian-genetics-known-plant (10), inclined-plane-friction-named-surfaces (10). 

Short tasks have oracle lengths less than 37 steps (median), and Long tasks have oracle lengths more than equal to 37 steps.

The map to the short names used for tasks in the paper:

Temp: use-thermometer, measure-melting-point-known-substance; Pick\&Place: find-plant, find-living-thing; Chemistry: chemistry-mix, chemistry-mix-paint-secondary-color; Lifespan: lifespan-longest-lived, lifespan-shortest-lived; Biology: identify-life-stages-1, identify-life-stages-2, Boil; Freeze; Grow Plant, Grow Fruit; Force: inclined-plane-determine-angle; Friction: inclined-plane-friction-named-surfaces; Genetics: mendelian-genetics-known-plant, mendelian-genetics-unknown-plant.

\begin{table}[t!]
\centering
{\small
\resizebox{0.6\linewidth}{!}{%
\begin{tabular}{@{}l@{}c||cc|cc@{}}
\toprule
  & & \multicolumn{2}{c|}{Generative L. Agents} & \multicolumn{2}{c}{\bf \clin{ (ours)}} \\
 \midrule
Task &  Type & ReAct & Reflexion & \textsc{Base} &  \textsc{Adapt} \\ \midrule
 Temp & S & 7.2 & 5.9 & 25.2 & 14.3 \\
 Temp & S & 6.1 & 28.6 & 53.2 & 51.8 \\
 Pick\&Place & S & 26.7 & 64.9 & 92.5 & 100.0 \\
 Pick\&Place & S & 53.3 & 16.4 & 55.0 & 100.0 \\
 Chemistry & S & 51.0 & 70.4 & 44.5 & 44.4 \\
 Chemistry & S & 58.9 & 70.7 & 56.7 & 56.7 \\
 Lifespan & S & 60.0 & 100.0 & 85.0 & 100.0 \\
 Lifespan & S & 67.5 & 84.4 & 70.0 & 90.0 \\
 Biology & S & 8.0 & 8.0 & 10.0 & 8.0 \\
 Boil & L & 3.5 & 4.2 & 7.0 & 15.2 \\
 Freeze & L & 7.8 & 7.8 & 10.0 & 10.0 \\
 GrowPlant & L & 9.1 & 7.3 & 10.2 & 11.1 \\
 GrowFruit & L & 18.6 & 13.0 & 35.9 & 71.6 \\
 Biology & L & 27.7 & 2.6 & 70.0 & 81.0 \\
 Force & L & 40.5 & 50.6 & 53.5 & 100.0 \\
 Friction & L & 44.0 & 100.0 & 56.5 & 72.5 \\
 Genetics & L & 25.7 & 50.9 & 77.4 & 100.0 \\
 Genetics & L & 16.8 & 23.7 & 62.3 & 92.6 \\
\midrule
 &  S & 37.6 & 49.9 & 54.7 & \bf 62.8 \\
 &  L & 21.5 & 28.9 & 42.5 & \bf 61.6 \\
 &  All & 29.6 & 39.4 & 48.6 & \bf 62.2 \\
\bottomrule
\end{tabular}%
}
}
\caption{Comparing \clin{} with baselines for \textbf{adaptation} \label{table:q1}}
\vspace{-3mm}
\end{table}

\end{document}